\documentclass[letterpaper,10pt,conference]{ieeeconf}
\IEEEoverridecommandlockouts
\usepackage{url}
\usepackage{graphics} % for pdf, bitmapped graphics files
\usepackage{float}
\usepackage{xcolor,amsmath, graphicx, float}
\usepackage[breakable, theorems, skins]{tcolorbox}
\usepackage{multirow}
\usepackage{makecell}
\usepackage{booktabs}
\let\labelindent\relax
\usepackage[inline, shortlabels]{enumitem}
\usepackage{color}
\usepackage{tcolorbox}
\usepackage{pifont}

\usepackage[hidelinks]{hyperref}

\usepackage{soul}
\newfloat{listing}{H}{lop}
\floatname{listing}{Listing}
 \setlist{nosep}  % removes extra space around lists

\def\BibTeX{{\rm B\kern-.05em{\sc i\kern-.025em b}\kern-.08em
    T\kern-.1667em\lower.7ex\hbox{E}\kern-.125emX}}
\begin{document}

% \title{Video Models are Frame by Frame Models \\At Least In Autonomous Driving
% }

% \title{On the Success and Failures of Video Models for Temporal Understanding \\We Are Not There Yet At Least In Autonomous Driving
% }

% \title{When Video Models are Frame-by-Frame in Autonomous Driving
% }

% \title{Mind the Spatio-Temporal Gap: On the Success and Failures of Video Models for Temporal Understanding in Driving}

\title{Mind the Temporal Gap: Evaluating Video Models for Temporal Understanding in Driving}

\author{\href{https://www.linkedin.com/in/shiva-sreeram-222b54250/}{ \color{violet}Shiva Sreeram$^1$},\quad \href{https://www.linkedin.com/in/erfan-aasi}{\color{violet}Erfan Aasi$^1$}, \quad \href{https://zswang666.github.io/}{\color{violet}Tsun-Hsuan Wang$^1$}, \quad
\href{https://alaamaalouf.github.io/}{\color{violet}Alaa Maalouf$^{2}$},\quad  and 
\href{https://www.csail.mit.edu/person/daniela-rus}{\color{violet}Daniela Rus$^1$}\\ \\
{\color{magenta}$^1$MIT CSAIL} \quad\quad {\color{magenta} $^{2}$University of Haifa }
\vspace{-5mm}
\thanks{
*Research was sponsored by the Department of the Air Force Artificial Intelligence Accelerator and was accomplished under Cooperative Agreement Number FA8750-19-2-1000. The views and conclusions contained in this document are those of the authors and should not be interpreted as representing the official policies, either expressed or implied, of the Department of the Air Force or the U.S. Government. The U.S. Government is authorized to reproduce and distribute reprints for Government purposes notwithstanding any copyright notation herein. \newline Author Alaa Maalouf thanks the Neubauer Family Foundation and the Council for Higher Education's MAOF program for their support. {$^1$MIT CSAIL}, { $^{2}$University of Haifa }.}
}

\maketitle

\begin{figure*}[t]
    \vspace{-5mm}
    \centering
    \includegraphics[width=0.865\linewidth]{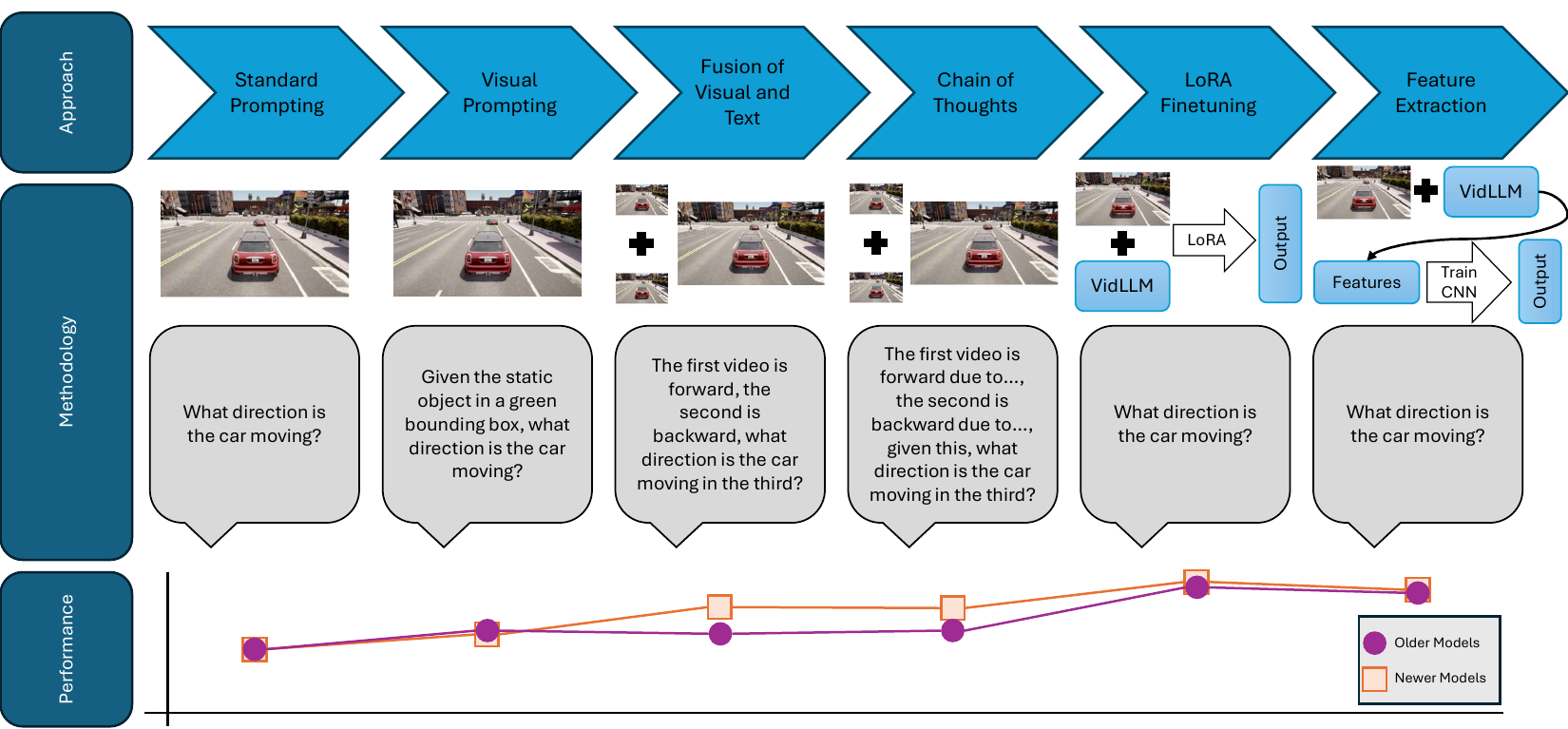}
    \caption{\textbf{Pipeline for probing ego-motion understanding in video-language models.} We evaluate increasingly structured approaches, from training-free prompting to lightweight learning-based probes. In \textit{standard prompting}, the model directly classifies the vehicle's motion. \textit{Visual prompting} adds a green bounding box around a static reference object to test whether explicit spatial grounding improves motion inference. In \textit{fusion of visual and text}, labeled few-shot video examples are provided before the query. \textit{Chain-of-thought prompting} explicitly relates temporal changes in the reference object's position or size to ego-motion. We then apply \textit{LoRA finetuning} to test whether motion distinctions can be learned under lightweight supervision, and \textit{feature extraction} to train simple classifiers on intermediate visual representations and examine where motion information remains decodable. Overall, prompt-level methods remain strongly model- and task-dependent, whereas learning-based approaches recover substantially stronger motion discrimination.}
    \vspace{-5.5mm}
    \label{fig:pipeline}
\end{figure*}

\begin{abstract}
Modern video-language models are increasingly treated as general-purpose video reasoners, yet it remains unclear whether they robustly capture temporal dynamics in safety-critical settings such as driving. We study this question through controlled ego-motion tasks that require identifying actions from driving videos. Across video-language models, we find that standard prompting, visual prompting, few-shot prompting, and chain-of-thought prompting remain unreliable and highly task- and model-dependent, with persistent class biases, although newer models show substantial task-specific few-shot gains. In contrast, lightweight LoRA finetuning and simple probes trained on intermediate visual features recover strong motion-discriminative performance, indicating that motion-relevant cues are often present inside the models but are not reliably used by the final video-to-language pathway. These results suggest that, in driving, current video-language models remain unreliable temporal reasoners and often behave more like frame-level, bias-driven recognizers than dependable temporal world models.
% \keywords{AI-enabled Robotics, Robot Vision, Robot Learning}

\end{abstract}
\newcommand{\vidlm}{VidLLM}
\vspace{-2mm}
\section{Introduction}
\vspace{-1mm}

%\section{Introduction}
\label{sec:intro}
\textbf{Video models} have undergone a rapid transformation in the last few years: from task-specific action recognizers to large-scale, spatiotemporal encoders trained with increasingly general objectives~\cite{zhu2020comprehensive,tong2022videomae,wang2022internvideo}. This evolution mirrors a central trend in modern machine learning: replacing narrowly engineered pipelines with large-scale models that can transfer across tasks, data regimes, and forms of supervision~\cite{bommasani2022opportunities,radford2021learning}. In the video domain, early progress established that attention-based architectures over space and time can rival, and in some cases outperform, classical 3D convolutional approaches for visual understanding~\cite{bertasius2021timesformer,arnab2021vivit}.  Later, self-supervised pretraining, made video representations more data-efficient and transferable \cite{tong2022videomae}, and more recent ``video foundation'' efforts aim to provide general-purpose spatiotemporal representations across diverse downstream tasks \cite{wang2022internvideo}. In parallel, aligning visual representations with language has produced \emph{large video-language models} (\vidlm{s}) that can caption, answer questions, and follow instructions over video inputs in a largely open-ended manner~\cite{tang2023vidllmsurvey,maaz2024videochatgpt,cheng2024videollama2,wang2024qwen2vl,bai2025qwen25vl,bai2025qwen3vl,wang2025internvl35,yu2025minicpmv45}, leading to a direct trend of video understanding inherited applications~\cite{suglia2024alanavlm,guan2024tasksuccess,sima2024drivelm,hwang2025emma}. 

%This was followed by advances in self-supervised pretraining, which significantly improved the efficiency and transferability of learned video representations~\cite{tong2022videomae}. Building on these developments, recent video foundation models aim to provide increasingly general-purpose spatiotemporal features for a wide variety of downstream applications~\cite{wang2022internvideo}. In parallel, the integration of language supervision has given rise to \emph{large video-language models} (\vidlm{s}), expanding the scope of video modeling from closed-set recognition toward open-ended captioning, question answering, and instruction following over temporally extended visual inputs~\cite{tang2023vidllmsurvey,maaz2024videochatgpt,cheng2024videollama2}.
%Early work demonstrated that attention over space and time can rival (and sometimes surpass) 3D convolutional pipelines for video understanding~\cite{bertasius2021timesformer,arnab2021vivit}.

This progress has fueled a natural hypothesis: \ul{if \vidlm{s} can describe what happens in a video, do they serve as \emph{world models}, i.e., internal predictive representations that capture how the world evolves over time}~\cite{ha2018worldmodels,hafner2023dreamerv3}.
Indeed, benchmark suites such as MVBench, Video-MME, and MMBench-Video report steady improvements and have become standard instruments for measuring video-language understanding \cite{li2024mvbench,fu2025videomme,fang2024mmbenchvideo}.
However, a missing (and practically consequential) question remains: \ul{\emph{do today’s \vidlm{s} genuinely reason about temporal dynamics, or do they behave primarily as strong frame-based models whose outputs are only weakly constrained by time?}}
Recent evidence suggests that commonly used video QA benchmarks can contain substantial spatial and textual shortcuts, where single frames, shuffled frames, or language priors can explain a surprising fraction of performance~\cite{cores2025tvbench}.
Concurrently, new benchmarks targeting spatio-temporal \emph{grounded} reasoning indicate that even strong models may fail to reliably align their reasoning with \emph{when} and \emph{where} evidence occurs~\cite{li2024mvbench,sugandhika2025knowshow,liu2024tempcompass}.

Autonomous driving makes this gap especially stark.
Driving scenes are fundamentally \emph{dynamic}: safe behavior depends on inferring ego-motion (forward/backward, left/right), as well as acceleration and deceleration, from temporally evolving cues.
Large-scale driving datasets (e.g., nuScenes and Waymo Open Dataset) therefore emphasize sequential, time-synchronized sensing precisely because motion and interaction are central to the task \cite{caesar2020nuscenes,sun2020waymo}.
Motivated by the promise of natural-language interfaces for driving agents, recent work has begun integrating vision-language models into driving-centric reasoning and interaction pipelines \cite{sima2024drivelm,drivelmm2025o1}.
Yet, despite this momentum, it is still unclear whether general-purpose \vidlm{s} can \emph{robustly} infer even basic ego-vehicle dynamics from videos---a prerequisite for treating them as reliable video reasoning engines in the loop.

\vspace{-2mm}
\subsection{Our Contributions}
\vspace{-1mm}
\textbf{This paper provides an experimental study of temporal motion reasoning in driving-focused video understanding.}
We design controlled prompts and evaluations that isolate basic ego-dynamics and ask whether \vidlm{s} can infer them from video segments. We then probe a spectrum of intervention levels---from basic prompting to adaptation (fine-tuning)---and complement behavioral evaluation with representation-level diagnostics. 
\begin{itemize}
    \item \textbf{Driving-focused temporal reasoning protocol.} We propose a targeted evaluation suite for ego-motion primitives (forward/backward, left/right) and longitudinal dynamics (acceleration/deceleration) from third-person car video.
    \item \textbf{Multi-level probing of \vidlm{s}.} We systematically test (1) standard prompting, (2) visually grounded/guided prompting, (3) text-vision fusion guided prompting, (4) chain-of-thoughts, (5) a combination of chain-of-thoughts with visual prompting, and (6) we further study lightweight parameter adaptation (e.g., LoRA) under comparable supervision. Finally, (7) we train small probing models on features extracted from different layers of the \vidlm{} as detailed next.
    \item \textbf{Where is motion information preserved or lost?} We probe intermediate visual features and find that motion cues may be preserved, strengthened, or degraded across architectures. Several models retain highly discriminative representations immediately before the language model despite poor prompted predictions, pointing to a downstream reasoning and generation bottleneck.
    
\end{itemize}
%\begin{figure}[t]
%\centering
%\begin{minipage}{0.49\linewidth}
%\centering
%\fbox{\parbox{0.95\linewidth}{\small
%\textbf{Example: confident but wrong.}\\
%(Frames thumbnail strip here)\\
%\emph{Model: ``The ego vehicle accelerates and turns left.''}\\
%\emph{GT: decelerates, straight.}
%}}
%\end{minipage}\hfill
%\begin{minipage}{0.49\linewidth}
%\centering
%\fbox{\parbox{0.95\linewidth}{\small
%\textbf{Fluency $\uparrow$ without correctness.}\\
%Tiny table/plot: Prompting level vs Accuracy\\
%(show near-chance accuracy across prompts)
%}}
%\end{minipage}
%\caption{Prompting increases narrative quality, but temporal motion accuracy remains near chance.}
%\label{fig:fluency_vs_correctness}
%\end{figure}

% Preamble:

% In the intro:
%\begin{tcolorbox}[title=\textbf{Key Findings (Preview)}, colback=black!2, colframe=black!25, arc=6pt]
%\begin{itemize}
%    \item \cmark\ Strong on \textbf{single-frame} perception \;\; \xmark\ Weak on \textbf{temporal ego-dynamics}
%    \item Prompts improve \textbf{fluency} \;(\emph{sounds right})\; but not \textbf{correctness} \;(\emph{is right})
%    \item LoRA/light fine-tuning: \textbf{no reliable jump} to motion reasoning
%    \item Probes find temporal signal \textbf{inside} the model, but it is \textbf{not used} at generation
%\end{itemize}
%\end{tcolorbox}

\noindent\textbf{A glimpse into our findings.}
Our main contribution is a striking set of empirical observations. Across the tested \vidlm{s}, we see a consistent pattern:

a. VidLLM ego-motion predictions remain \ul{highly model- and task-dependent under prompting}, with persistent class biases and \emph{poor forward/backward discrimination}.

b. \ul{More elaborate/sophisticated prompting} including structured guided templates and step-by-step reasoning instructions, typically improves \emph{fluency} (outputs become more coherent and confident) but \ul{does not universally improve \emph{temporal correctness}}, where some newer models can be elicited into much better performance, but the gain fails to transfer consistently across ego-motion categories.

c. Adaptation can unlock motion reasoning:
\ul{lightweight LoRA fine-tuning can reliably turn these models into robust motion reasoners}, suggesting that the underlying architectures \emph{can} represent temporal cues and that a key bottleneck is the availability/quality of supervision (used data and training procedure) for dynamics.

d. Finally, layer-wise probing shows that motion information remains strongly decodable before the language model, yet is not reliably reflected in final predictions, indicating a gap between visual representation and language-based reasoning.

Together, these caution against treating off-the-shelf \vidlm{s} as reliable video world models for driving without explicit temporal grounding and dynamics-aware training.

%\noindent\textbf{Video Models.}

%\noindent\textbf{video models in driving. ... mention what is missing.}"raise a question"

%\textbf{Our contribution}

% \section{EXPERIMENTAL STUDY}

\vspace{-2mm}
\section{Probing via Different Levels.}
\vspace{-1mm}
\label{sec:probing}
%This work aims to show the current capabilities of state-of-the-art (SOTA) video models, namely VideoLLaMA2 and Qwen-2-VL, in identifying and reasoning about motions valuable to understanding driving scenarios and more importantly, a variety of steps taken to boost performance and/or reasoning.

% \section{METHODS}

%%%% Show how these look under each of the cases. Under each there is a running example of what this looks like. Under each have a simple plot of x-axis being "difficulty" and y-axis being accuracy
%Our goal is to probe and obtain the models' capabilities in reasoning through the ego motion driving tasks. We begin with prompting the model to ask the standard query on what motion is followed, transition to applying a set of prompt level techniques, and then explore learning approaches; all of these have the goal of identifying the ways in which the video models can or cannot reason about vehicular motions.
We propose a multi-level probing framework to analyze the extent to which video-language models (\vidlm{s}) can reason about ego-motion in driving settings. Our approach systematically varies the level of intervention applied to the model. At the lowest level, we use standard prompting to measure baseline capabilities in identifying and describing motion patterns. At an intermediate level, we apply targeted prompt-based techniques to elicit temporally consistent and causally grounded reasoning, testing whether deficiencies arise from inference procedures rather than representation. At the highest level, we incorporate lightweight learning-based adaptations, enabling the model to refine its predictions and revealing whether its limitations are fundamental or correctable. By comparing performance across these levels, we isolate where reasoning breaks down and characterize both the latent strengths and the critical gaps of \vidlm{s} in modeling vehicular dynamics.

\vspace{-1.5mm}
\subsection{Prompt-Level Probing Strategies}
\vspace{-1mm}
\label{subsec:prompt-probing}
\noindent\textbf{(1) Standard prompt probing. } We provide a set of binary questions depending on the scenario where we simply query the model what type of motion is being taken by the car in the video. This follows prompting strategies in related probing works \cite{sreeram2025probing} to obtain the results.

\noindent\textbf{(2) Visual prompting}: Here we aim to guide the models' reasoning by choosing a static object in every video and placing a green bounding box. The objective is to extend task prompting to the image domain which has shown success in foreground segmentation and single object detection \cite{NEURIPS2022_9f09f316}. This box is placed via open-set detection, tracking, and following \cite{maalouf2024follow}. This test reveals whether highlighting a stable spatial anchor helps the model correctly infer motion by focusing its attention on consistent visual reference points.\\
% Expand this to include more "latex" terms, frame $i$, mask at pixel values

\noindent\textbf{(3) Fusion of visual and text}: Here we aim to improve the models' capabilities by providing an example of two types of motions (i.e. forward and backward) before providing the final video. This is an application of few-shot learning, which aims to generalize reasoning from a few examples \cite{wang2020generalizing}. Here, we evaluate whether the model can internalize simple motion patterns from a small number of demonstrations and generalize them to unseen videos.\\

\noindent\textbf{(4) Chain-of-thoughts}: Here, we aim to guide the model through the scenario such that we ensure the model understands how changes in the background can be reasoned through via chain-of-thoughts \cite{wei2022chain}. This is combined with the few-shot approach where it is explained to the model the way in which the cases can be identified to be a specific type of motion. In our videos, we would observe the way in which the static object shifting through the scene can lead to the correct interpretation of the motion. For example, the size of a static object in the background increasing means that the object is getting closer, which means that the ego is moving forward. This test assesses whether explicit step-by-step reasoning enables the model to connect background changes to vehicular motion, thereby compensating for missing implicit temporal understanding.

\noindent\textbf{(5) Chain-of-thoughts with visual prompting}: This aims to combine all of the prompt level strategies in order to provide the best reasoning capabilities to the model. In essence, the visual prompting approach of surrounding a static object with a bounding box is provided, with chain-of-thoughts reasoning to guide the model in understanding the way in which the bounding box shifts enables understanding of the motion. For example, the size of the bounding box in the background increasing means that the static object is getting closer, which means that the ego is moving forward. Through this probing technique, we examine whether combining structured reasoning with a visual reference object yields synergistic improvements in motion understanding beyond what either cue provides alone.

\vspace{-2mm}
\subsection{Probing via Learning Techniques}
\vspace{-1mm}
Here we aim to test the models capabilities through training techniques. This process can harm the models' generalization but we can see how it affects performance.\\

\noindent\textbf{(6) LoRA finetuning}: To evaluate whether the video-language models possess latent representational structure that can be unlocked through lightweight adaptation, we apply Low-Rank Adaptation (LoRA) \cite{hu2022lora} to finetune the model on our driving-motion dataset. LoRA introduces trainable low-rank matrices into the attention and projection layers while freezing the original model weights, enabling efficient specialization without altering the underlying pretrained parameters. During training, each video is paired with its ground-truth label, and the LoRA modules update rank-$r$ adapters on top of pretrained weights.

\noindent\textbf{(7) Feature extraction}: To further understand where ego-motion information is represented within the models' visual processing pipelines, we extract features from two locations in each architecture: immediately before and immediately after the model-specific visual compression or projection stage. We then train simple classifiers on top of these frozen representations. This allows us to test whether motion cues are already encoded by the visual backbone, whether they survive the compression into the representation consumed by the language model, and whether failures under prompting arise despite the presence of decodable motion information. We evaluate several lightweight probes, including: (i) one linear layer, (ii) one linear layer with pooling, (iii) a one-layer CNN, and (iv) a one-layer CNN with pooling.

\noindent\textbf{(7a) Pre-Compression Visual Features:}
In the first setting, we extract the visual representations immediately before the architecture-specific module that compresses, resamples, merges, or projects the visual token sequence toward the language model. These representations therefore retain the higher-resolution visual token structure produced by the vision backbone. Importantly, we do not assume that these representations are purely spatial or devoid of temporal information. Depending on the architecture, temporal context may already have been incorporated by the visual encoder.

\begin{itemize}
    \item \textit{VideoLLaMA2}~\cite{cheng2024videollama2}: We extract the output of the ViT-based vision tower immediately before the Spatio-Temporal Connector (STC). These features retain the per-frame visual representations prior to the explicit connector that compresses and reorganizes the visual sequence for the language model.

    \item \textit{Qwen2-VL-7B-Instruct}~\cite{wang2024qwen2vl}: We extract the output of the visual transformer immediately before the model's visual token merger. At this stage, the representation retains the higher-resolution visual token grid produced by the vision encoder, prior to token compression and projection into the language-model embedding space.

    \item \textit{Qwen2.5-VL-7B-Instruct}~\cite{bai2025qwen25vl}: We analogously extract the output of the visual transformer before the visual PatchMerger. These features preserve the higher-resolution spatiotemporal token representation before neighboring visual tokens are merged and projected to the dimensionality expected by the language model.

    \item \textit{Qwen3-VL-8B-Instruct}~\cite{bai2025qwen3vl}: We extract visual-transformer features immediately before the model's visual merger. As in the earlier Qwen architectures, this representation retains a denser visual token grid prior to spatial token compression and projection. The temporal structure of the video remains represented at this stage, so these features should not be interpreted as purely frame-independent spatial encodings.

    \item \textit{InternVL3.5-8B}~\cite{wang2025internvl35}: We extract the visual features prior to the model's downsampling and projection pathway that converts the high-resolution vision-backbone representation into the more compact sequence provided to the language model. These features therefore represent the higher-resolution output of the visual encoder before cross-modal compression.

    \item \textit{MiniCPM-V 4.5-8B}~\cite{yu2025minicpmv45}: We extract the output of the vision tower before the model's resampler. At this point, the representation consists of the higher-resolution visual features produced by the vision encoder, before they are compressed into the fixed latent visual representation consumed by the language model.
\end{itemize}

These pre-compression representations allow us to test whether ego-motion cues are already recoverable from the visual backbone before the model reduces or restructures the visual token sequence.

\noindent\textbf{(7b) Post-Compression / Projected Visual Features:}
In the second setting, we extract features after the model-specific visual compression, resampling, or projection operation. These features are substantially closer to the representation actually supplied to the language model and therefore allow us to determine whether motion information survives the visual-to-language interface. The precise operation differs: it may correspond to an explicit spatiotemporal connector, a spatial token merger and projection layer, a visual downsampling/projector pathway, or a learned resampler.

\begin{itemize}
    \item \textit{VideoLLaMA2}: We extract features after the Spatio-Temporal Connector (STC). The STC compresses and reorganizes the visual sequence while explicitly incorporating spatiotemporal structure before the resulting tokens are passed to the language model.

    \item \textit{Qwen2-VL-7B-Instruct}: We extract the output after the visual token merger and projection stack. The merger reduces the number of visual tokens by combining neighboring visual representations and projects the resulting tokens into the language representation.

    \item \textit{Qwen2.5-VL-7B-Instruct}: We extract features after the visual PatchMerger and its associated projection. Relative to the pre-merge representation, this stage contains fewer visual tokens with a larger projected feature dimension and corresponds more closely to the visual representation presented to the language model.

    \item \textit{Qwen3-VL-8B-Instruct}: We similarly extract the representation after the visual merger and projection stage. For example, the merger spatially compresses the visual token grid while retaining its temporal organization, producing a smaller set of higher-dimensional tokens for subsequent language-model processing.

    \item \textit{InternVL3.5-8B}: We extract the representation after the model's visual downsampling and projection pathway. This representation is the compressed visual-token sequence produced from the vision-backbone features before those tokens interact with the language model.

    \item \textit{MiniCPM-V 4.5-8B}: We extract features after the visual resampler. The resampler transforms the higher-resolution vision-tower output into a compact set of latent visual tokens that constitute the representation supplied to the language model. For CNN-based probing, these latent tokens are reshaped into a regular feature layout for classification. This layout is used only for the probe and should not be interpreted as a native spatial grid of the MiniCPM architecture.
\end{itemize}

Comparing the pre- and post-compression probes allows us to distinguish between two possible failure modes. If motion information is recoverable before compression but lost afterward, the visual-to-language compression stage itself may be discarding task-relevant information. Conversely, if ego-motion remains highly decodable after compression but the model still fails under natural-language prompting, this suggests that the information survives into the representation available to the language model but is not reliably accessed or converted into the correct linguistic prediction.

%%%% explain better why we used two kind of features!
%%%% perhaps 6a and 6b for the type of features we pull from the model
\vspace{-3mm}
\section{Experimental Results}
\vspace{-1.5mm}

\begin{table*}[t]
\vspace{-4mm}
\centering
\scriptsize
\setlength{\tabcolsep}{3pt}
\caption{Accuracy on ego-motion tasks across video-language models under different prompting and learning-based strategies. Standard prompting is evaluated on both simulated (Sim) and real-world (Real) driving videos; all other results use Sim data. Features (7a) and (7b) report the 1-layer CNN probe without pooling.}
\resizebox{\textwidth}{!}{%
\begin{tabular}{c || c || c c c c c c || c c c}
\toprule
\multirow{3}{*}{Model}
    & \multirow{3}{*}{Action Pair}
    & \multicolumn{6}{c||}{Prompt-level probing (accuracy \%)}
    & \multicolumn{3}{c}{Learning-based probing (accuracy \%)} \\

& 
    & \multicolumn{2}{c}{Standard}
    & \multirow{2}{*}{Visual Prompt (VP)}
    & \multirow{2}{*}{Few-shot}
    & \multirow{2}{*}{Few-shot + CoT}
    & \multirow{2}{*}{Few-shot + CoT + VP}
    & \multirow{2}{*}{LoRA Finetune}
    & \multirow{2}{*}{Features (7a)}
    & \multirow{2}{*}{Features (7b)} \\

& & Sim & Real & & & & & & & \\
\midrule

\multirow{3}{*}{Qwen2-VL}
    & Forward / Backward      & 50.00 & 50.00 & 65.00 & 58.80 & 62.06 & 62.06 & 100.00 & 88.75 & 93.75 \\
    & Left / Right            & 50.00 & 50.00 & 50.00 & 48.57 & 50.00 & 49.27 & 96.25  & 93.75 & 97.50 \\
    & Accelerate / Decelerate & 51.25 & 35.00 & 51.00 & 50.00 & 50.00 & 48.50 & 100.00 & 87.50 & 88.75 \\
\midrule

\multirow{3}{*}{Qwen2.5-VL}
    & Forward / Backward      & 50.00 & 50.00 & 54.27 & 50.00 & 50.75 & 51.26 & 100.00 & 50.00 & 97.50 \\
    & Left / Right            & 50.00 & 50.00 & 50.00 & 50.00 & 50.00 & 50.00 & 100.00 & 77.50 & 100.00 \\
    & Accelerate / Decelerate & 51.50 & 55.00 & 51.00 & 51.26 & 50.25 & 50.25 & 100.00 & 83.75 & 91.25 \\
\midrule

\multirow{3}{*}{Qwen3-VL}
    & Forward / Backward      & 50.00 & 50.00 & 50.00 & 50.25 & 54.02 & 52.76 & 100.00 & 55.00 & 97.50 \\
    & Left / Right            & 60.00 & 85.00 & 59.00 & 72.61 & 60.80 & 59.05 & 100.00 & 52.50 & 100.00 \\
    & Accelerate / Decelerate & 56.00 & 50.00 & 57.00 & 77.14 & 68.84 & 68.09 & 100.00 & 83.75 & 85.00 \\

\midrule

\multirow{3}{*}{InternVL 3.5}
    & Forward / Backward      & 50.00 & 50.00 & 50.00 & 50.25 & 50.25 & 50.00 & 100.00 & 100.00 & 65.00 \\
    & Left / Right            & 67.00 & 75.00 & 63.50 & 70.10 & 56.78 & 54.77 & 100.00 & 100.00 & 100.00 \\
    & Accelerate / Decelerate & 54.50 & 50.00 & 52.00 & 53.02 & 56.28 & 54.02 & 100.00 & 98.75 & 93.75 \\
\midrule

\multirow{3}{*}{MiniCPM-V 4.5}
    & Forward / Backward      & 50.00 & 50.00 & 50.00 & 50.00 & 50.00 & 50.00 & 100.00 & 67.50 & 100.00 \\
    & Left / Right            & 49.50 & 50.00 & 50.00 & 51.76 & 50.00 & 50.00 & 100.00 & 100.00 & 100.00 \\
    & Accelerate / Decelerate & 50.00 & 50.00 & 50.50 & 55.78 & 50.25 & 50.00 & 97.50 & 97.50 & 97.50 \\

\bottomrule
\end{tabular}}

\label{tab:all_models_strategies}

\end{table*}

% \begin{table}[t]
%     \centering
%     \vspace{4mm}
%     \footnotesize
%     \caption{VideoLLaMA2 accuracy on ego-motion tasks under different prompting and learning-based strategies.}
%     \vspace{-1mm}
%    \resizebox{\columnwidth}{!}{
%     \begin{tabular}{c || c c | c c c}
%     \toprule
%     \multirow{2}{*}{Action Pair} 
%         & \multicolumn{2}{c|}{Prompt-level probing (accuracy \%)} 
%         & \multicolumn{3}{c}{Learning-based probing (accuracy \%)} \\
%     & Standard & Visual prompt 
%     & LoRA finetune & Features (7a) & Features (7b) \\
%     \midrule
%     Forward / Backward      & 50.00 & 50.00 & 100.0 & 98.75 & 77.50 \\
%     Left / Right            & 50.50 & 49.50 & 97.50 & 100.0 & 100.0 \\
%     Accelerate / Decelerate & 49.75 & 50.00 & 82.50 & 86.25 & 66.25 \\
%     \bottomrule
%     \end{tabular}}
%     \label{tab:videollama2_strategies}
% \end{table}

\begin{table}[t]
    \centering
    \vspace{-4mm}
    \footnotesize
    \caption{VideoLLaMA2 ego-motion accuracy. Standard prompting uses simulated (Sim) and real-world (Real) videos; all other results use Sim data. Feature columns follow Table~\ref{tab:all_models_strategies}.}
    \vspace{-1mm}
    \resizebox{\columnwidth}{!}{
    \begin{tabular}{c || c c c | c c c}
    \toprule
    \multirow{3}{*}{Action Pair}
        & \multicolumn{3}{c|}{Prompt-level probing (accuracy \%)}
        & \multicolumn{3}{c}{Learning-based probing (accuracy \%)} \\
    
    & \multicolumn{2}{c}{Standard}
        & \multirow{2}{*}{Visual Prompt}
        & \multirow{2}{*}{LoRA Finetune}
        & \multirow{2}{*}{Features (7a)}
        & \multirow{2}{*}{Features (7b)} \\
    
    & Sim & Real & & & & \\
    \midrule
    
    Forward / Backward
        & 50.00 & 50.00 & 50.00 & 100.00 & 98.75 & 77.50 \\
    
    Left / Right
        & 50.50 & 50.00 & 49.50 & 97.50 & 100.00 & 100.00 \\
    
    Accelerate / Decelerate
        & 49.75 & 70.00 & 50.00 & 82.50 & 86.25 & 66.25 \\
    
    \bottomrule
    \end{tabular}}
    
    \label{tab:videollama2_strategies}
    
\end{table}

\paragraph{Overview.}
We evaluate whether modern video-language models (VidLLMs) can reliably infer ego-motion from driving videos. Across all experiments, we observe a consistent pattern: \textbf{models fail under prompting despite having access to sufficient information internally}. We support this claim through three complementary analyses: (i) prompt-level probing, (ii) lightweight finetuning, and (iii) representation-level probing.

\textbf{Video data.}
Our videos consist of CARLA scenarios in a third-person-perspective of the ego vehicle driving through a scene, following the given motion. As a ``reference'' point for the model to reason with, we add another static object (typically another vehicle) that the model can focus on. In total, we have 200 videos of every motion type: as in 200 forward videos, 200 backward videos, etc. The questions posed to the models are of a binary variety (i.e. is the car driving forward or backward?) with informative text dependent on the prompting strategy.

Figure~\ref{fig:pipeline} summarizes our probing pipeline, which spans both training-free prompt-level interventions and lightweight learning-based evaluations. The prompt-level methods modify how the video is presented to the frozen model, while the learning-based methods test whether motion-relevant information can be extracted or adapted from the model's internal representations.

\textbf{Real-world video data.}
To evaluate whether the observed behaviors extend beyond simulation, we additionally collect 60 real-world driving videos from a first-person/in-car ego perspective, with 10 examples for each of the six motion classes. For each motion class, half of the videos are recorded during the day and half at night, introducing variation in illumination and visual conditions. Because this dataset is substantially smaller than the simulated dataset, its reported accuracies are inherently less stable. The limited number of examples also makes it unsuitable for the supervised LoRA and feature-probing experiments, so we use the real-world data only for standard prompting.

\noindent\textbf{Models evaluated.}
We evaluate six video-language models spanning several recent open-source
architectures: VideoLLaMA2~\cite{cheng2024videollama2},
Qwen2-VL~\cite{wang2024qwen2vl},
Qwen2.5-VL~\cite{bai2025qwen25vl},
Qwen3-VL~\cite{bai2025qwen3vl},
InternVL 3.5~\cite{wang2025internvl35}, and
MiniCPM-V 4.5~\cite{yu2025minicpmv45}.

\noindent\textbf{Evaluation protocol.}
The simulated dataset contains 1,200 CARLA videos at $1280\times720$ and 60 FPS, while the real-world set contains 60 videos at $1920\times1080$ and 25 FPS. Clips longer than 10 seconds are truncated before inference. For prompted inference, VideoLLaMA2, InternVL 3.5, and MiniCPM-V 4.5 use 8 sampled frames per 10-second clip, while Qwen2-VL, Qwen2.5-VL, and Qwen3-VL sample at 2 FPS, yielding approximately 20 frames for a 10-second clip. Simulated standard/visual prompting uses all 200 videos per class; few-shot uses one demonstration per class and evaluates the remaining 199. Real-world evaluation uses standard prompting on all 10 videos per class. LoRA and feature probes use an 80/20 train/held-out split for each 400-video motion pair, yielding 80 held-out clips per pair (for example, 97.5\% corresponds to 78/80 correct predictions).

Here we showcase the motion understanding results:

\vspace{-3mm}
\subsection{Standard Prompting}
\vspace{-1.5mm}

We first evaluate each model using a basic binary prompt describing the static reference vehicle. Table~\ref{tab:base_model_motion_accuracies} reveals strong class biases rather than simple uncertainty. Most notably, \textbf{all models achieve 100\% accuracy on forward motion and 0\% on backward motion}, completely collapsing to the forward class, consistent with~\cite{sreeram2025probing}. Similar model-dependent biases appear for the other motion pairs: Qwen2-VL and Qwen2.5-VL strongly favor leftward motion, Qwen3-VL and InternVL 3.5 favor acceleration and rightward motion, and MiniCPM-V 4.5 predicts deceleration perfectly while failing on acceleration. Consequently, aggregate accuracy can obscure severe imbalance; for example, InternVL 3.5 reaches 67\% on left/right despite class accuracies of only 39\% and 95\%. Thus, the near-chance results in Tables~\ref{tab:all_models_strategies} and~\ref{tab:videollama2_strategies} primarily reflect systematic prediction biases rather than reliable use of temporal ego-motion cues.

\begin{figure}[t]
    \centering
    \vspace{-2mm}
    \includegraphics[width=\columnwidth]{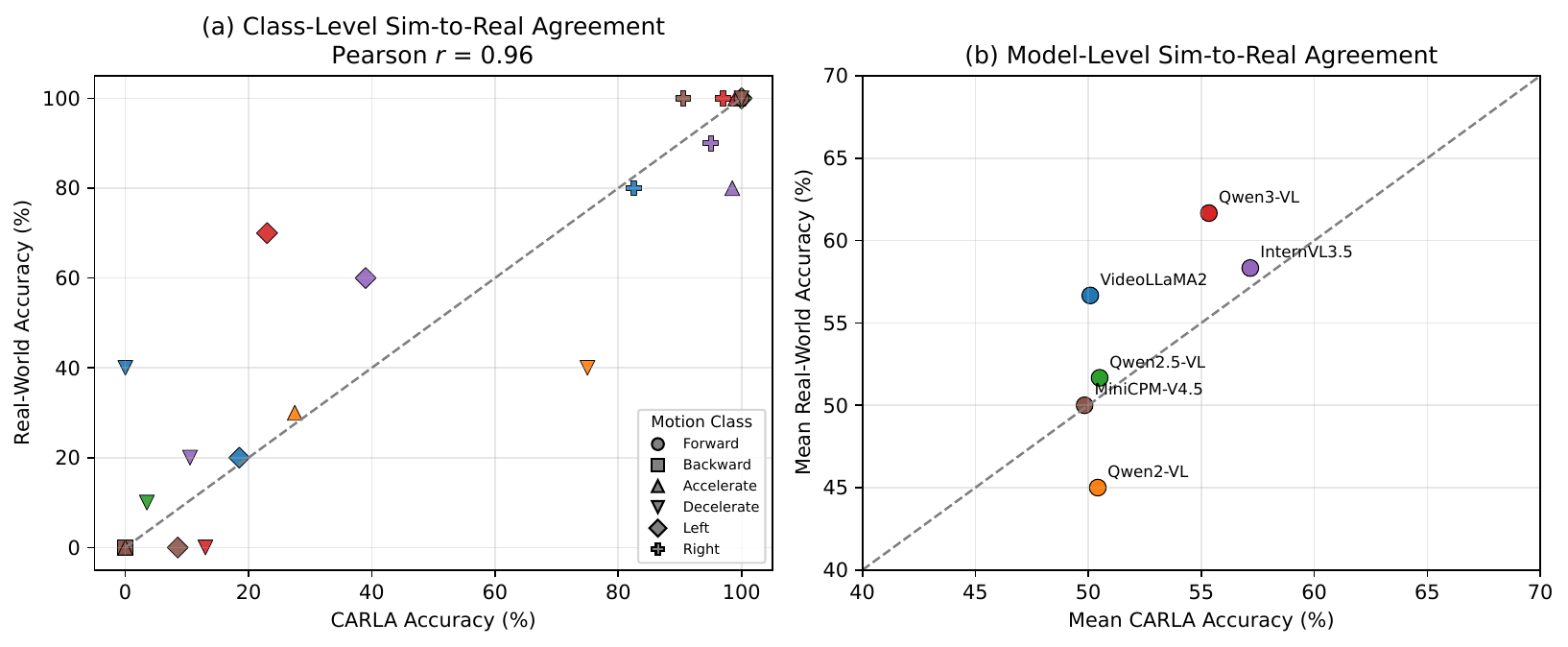}
    \caption{Sim-to-real comparison of standard-prompting accuracy. The same class-level biases appear across CARLA and real-world videos. Overall class-level correlation is $r=0.957$.}
    \label{fig:sim_real_comparison}
\end{figure}

\textbf{Sim-to-real consistency.}
Figure~\ref{fig:sim_real_comparison} shows many of the same class biases in real-world video. Although the class-level correlation is $r=0.957$, this value is inflated by the forward/backward collapse across models and should not be interpreted as an independent measure of sim-to-real generalization. Given the small set, we treat this primarily as evidence that the same biases appear outside CARLA.

\begin{table}[t]
    \centering
    \vspace{-4mm}
    \caption{Standard prompting accuracies for simulated data.}
    \scalebox{0.65}{
    \begin{tabular}{c || c c c c c c}
    \toprule
    Action & VideoLLaMA2 & Qwen2-VL & Qwen2.5-VL & Qwen3-VL & Intern-VL-3.5 & MiniCPM-v4.5 \\
    \midrule
    Forward     & 100.0 & 100.0 & 100.0 & 100.0 & 100.0 & 100.0 \\
    Backward    & 0.00   & 0.00 & 0.00 & 0.00 & 0.00 & 0.00 \\
    Accelerate  & 99.50  & 27.50 & 99.50 & 99.00 & 98.50 & 0.00 \\
    Decelerate  & 0.00   & 75.00 & 3.50 & 13.00 & 10.50 & 100.0 \\
    Left        & 18.50  & 100.0 & 100.0 & 23.00 & 39.00 & 8.50 \\
    Right       & 82.50  & 0.00 & 0.00 & 97.00 & 95.00 & 90.50 \\
    \bottomrule
    \end{tabular}}
    \label{tab:base_model_motion_accuracies}
\end{table}

% \begin{table}[t]
%     \centering
%     \caption{Base model accuracies for predicting ego-motion on real-world driving videos with no assistance. The real-world evaluation contains 60 videos, with 10 videos for each of the six motion classes.}
%     \scalebox{0.65}{
%     \begin{tabular}{c || c c c c c c}
%     \toprule
%     Action & VideoLLaMA2 & Qwen2-VL & Qwen2.5-VL & Qwen3-VL & Intern-VL-3.5 & MiniCPM-v4.5 \\
%     \midrule
%     Forward     & 100.0 & 100.0 & 100.0 & 100.0 & 100.0 & 100.0 \\
%     Backward    & 0.00  & 0.00  & 0.00  & 0.00  & 0.00  & 0.00 \\
%     Accelerate  & 100.0 & 30.00 & 100.0 & 100.0 & 80.00 & 0.00 \\
%     Decelerate  & 40.00 & 40.00 & 10.00 & 0.00  & 20.00 & 100.0 \\
%     Left        & 20.00 & 100.0 & 100.0 & 70.00 & 60.00 & 0.00 \\
%     Right       & 80.00 & 0.00  & 0.00  & 100.0 & 90.00 & 100.0 \\
%     \bottomrule
%     \end{tabular}}
%     \label{tab:real_base_model_motion_accuracies}
% \end{table} 

\vspace{-2mm}
\subsection{Visual Prompting}
\vspace{-1mm}
We next test whether explicitly grounding the model to a static reference object improves ego-motion reasoning by placing a green bounding box around a static vehicle and identifying it in the prompt. As shown in Tables~\ref{tab:all_models_strategies} and~\ref{tab:videollama2_strategies}, visual prompting provides little benefit. The only notable improvement is for Qwen2-VL on forward/backward motion, increasing from 50.00\% to 65.00\%.

\textbf{Thus, visual grounding alone is insufficient.} Providing a salient reference object does not reliably help the models use its temporal evolution across frames, suggesting that the limitation lies beyond simple spatial anchoring.

\vspace{-2mm}
\subsection{Fusion of Visual and Text}
\vspace{-1mm}
We next test whether few-shot visual-text prompting can reduce the class biases observed under standard prompting. The model is shown one labeled example from each class in the motion pair and then classifies a query video, leaving 199 test examples per motion class. Since few-shot results use 199 query clips per class, corresponding to a binomial standard error of approximately 3.5 percentage points near chance; differences of only a few points should therefore not be over-interpreted.

As shown in Table~\ref{tab:all_models_strategies}, few-shot performance is highly model- and task-dependent. Qwen2-VL improves only on forward/backward, from 50.00\% to 58.80\%, while Qwen2.5-VL remains near chance. In contrast, Qwen3-VL reaches 72.61\% on left/right and 77.14\% on accelerate/decelerate, and InternVL 3.5 reaches 70.10\% on left/right. However, forward/backward remains near chance. Thus, \textbf{few-shot demonstrations can improve motion discrimination for selected models and tasks, but do not provide consistent temporal reasoning across motion categories}.

VideoLLaMA2 does not support the same few-shot setup, so we approximate it by concatenating demonstration and query videos. Because this out-of-distribution format causes severe class collapse, we omit these results from Table~\ref{tab:videollama2_strategies}.

\begin{figure}[t]
    \vspace{-4mm}
    \centering
    \includegraphics[width=0.85\columnwidth]{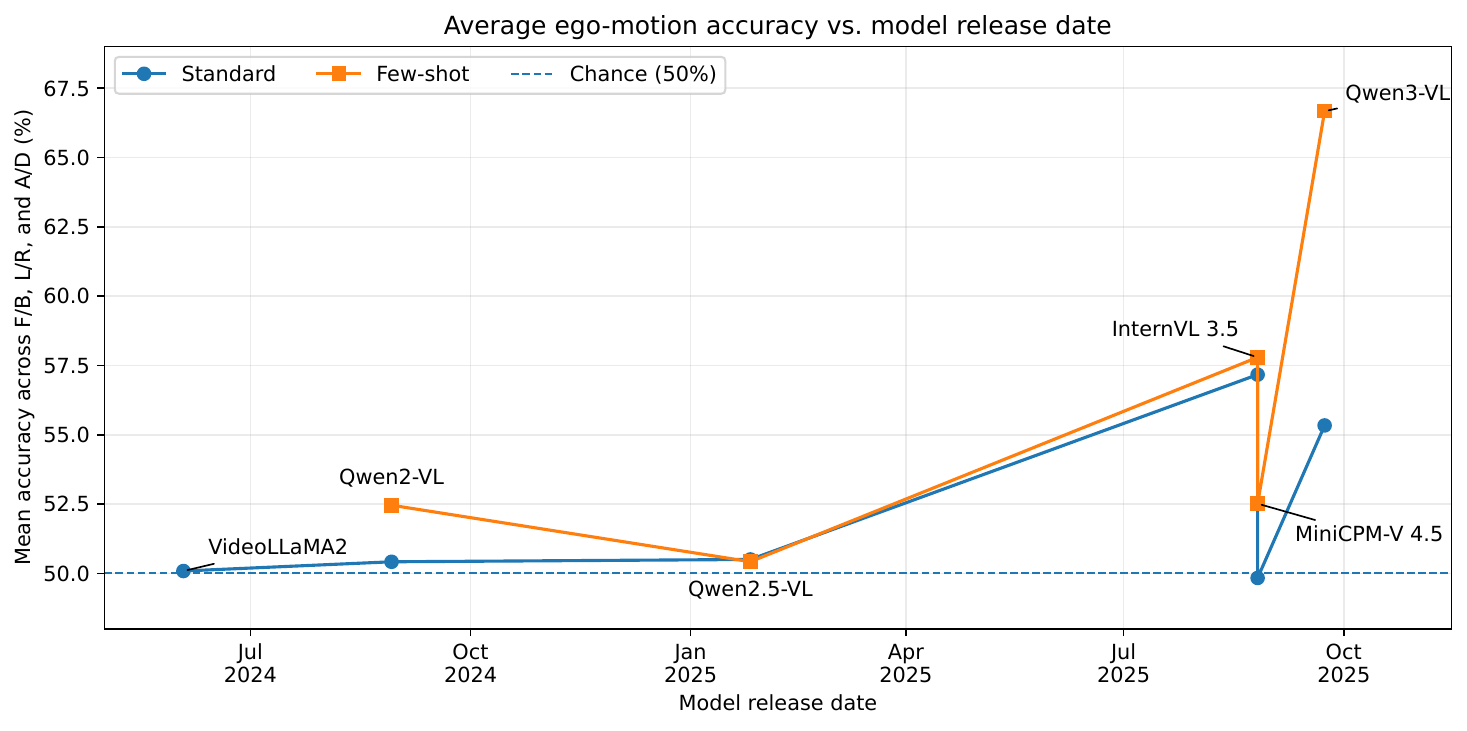}
    \vspace{-2mm}
    \caption{Average ego-motion accuracy versus model release date. Standard and few-shot results are averaged across motion pairs. VideoLLaMA2 is excluded from the few-shot trend because its protocol is not comparable.}
    \label{fig:release_date_performance}
\end{figure}

Figure~\ref{fig:release_date_performance} examines whether ego-motion performance improves across model generations. Standard prompting shows only modest gains over time, while few-shot prompting improves more substantially, particularly Qwen3-VL. However, these gains are task-dependent: forward/backward accuracy remains near chance, with most improvements coming from left/right and accelerate/decelerate reasoning.

\vspace{-2mm}
\subsection{Chain-of-Thoughts}
\vspace{-1mm}

We next evaluate whether chain-of-thought (CoT) prompting improves few-shot ego-motion reasoning by explicitly guiding the model through intermediate temporal cues. As shown in Table~\ref{tab:all_models_strategies}, CoT provides no consistent benefit. Qwen2-VL remains strongly class-biased, while Qwen2.5-VL stays near chance across all three motion pairs. Qwen3-VL reaches 60.80\% on left/right and 68.84\% on accelerate/decelerate, but both are worse than its plain few-shot results of 72.61\% and 77.14\%. InternVL 3.5 similarly drops from 70.10\% to 56.78\% on left/right, and MiniCPM-V 4.5 shows no consistent gain. Combining visual prompting with few-shot or CoT prompting also yields no consistent gains.

\textbf{Thus, CoT does not reliably improve temporal reasoning correctness.}

% \begin{table}[t]
%     \centering
%     \footnotesize
%     \caption{Comparison between few-shot methods for Qwen2-VL-7B-Instruct.}
%     \resizebox{\columnwidth}{!}{%
%     \begin{tabular}{c || c c c c}
%     \toprule
%     Action 
%         & Few shot 
%         & Few shot with CoT 
%         & Few shot with VP 
%         & Few shot with CoT and VP \\
%     \midrule
%     Forward     & 95.48 & 91.96 & 94.96 & 91.96 \\
%     Backward    & 22.11 & 32.16 & 21.16 & 32.16 \\
%     Accelerate  & 100.0 & 100.0 & 100.0 & 96.98 \\
%     Decelerate  & 0.00  & 0.00  & 0.00  & 0.02 \\
%     Left        & 5.53  & 0.00  & 5.03  & 8.04 \\
%     Right       & 91.60 & 100.0 & 98.5  & 90.5 \\
%     \bottomrule
%     \end{tabular}}
%     \label{tab:qwen_fewshot_methods}
% \end{table}

\begin{table}[t]
    \vspace{-4mm}
    \centering
    \footnotesize
    \caption{Comparison of few-shot prompting strategies across video-language models. VP denotes visual prompting and CoT denotes chain-of-thought prompting.}
    \resizebox{0.85\columnwidth}{!}{%
    \begin{tabular}{c || c || c c c c}
    \toprule
    Model
        & Action
        & Few-shot
        & Few-shot with CoT
        & Few-shot with VP
        & Few-shot with CoT and VP \\
    \midrule

    \multirow{6}{*}{Qwen2-VL}
        & Forward     & 95.48 & 91.96 & 94.96 & 91.96 \\
        & Backward    & 22.11 & 32.16 & 21.16 & 32.16 \\
        & Accelerate  & 100.0 & 100.0 & 100.0 & 96.98 \\
        & Decelerate  & 0.00  & 0.00  & 0.00  & 0.02 \\
        & Left        & 5.53  & 0.00  & 5.03  & 8.04 \\
        & Right       & 91.60 & 100.0 & 98.50 & 90.50 \\

    \midrule

    \multirow{6}{*}{Qwen2.5-VL}
        & Forward     & 100.00 & 100.00 & 97.99 & 99.50 \\
        & Backward    & 0.00   & 1.51   & 10.55 & 3.02 \\
        & Accelerate  & 99.50  & 38.69  & 98.99 & 40.20 \\
        & Decelerate  & 3.02   & 61.81  & 5.03 & 60.30 \\
        & Left        & 100.00 & 100.00 & 100.00 & 100.00 \\
        & Right       & 0.00   & 0.00   & 0.00 & 0.00 \\

    \midrule

    \multirow{6}{*}{Qwen3-VL}
        & Forward     & 100.00 & 97.99 & 100.00 & 100.00 \\
        & Backward    & 0.50   & 10.05 & 0.00   & 5.53 \\
        & Accelerate  & 85.43  & 91.96 & 89.45 & 92.46 \\
        & Decelerate  & 68.84  & 45.73 & 64.32 & 43.72 \\
        & Left        & 71.36  & 63.32 & 69.85 & 62.31 \\
        & Right       & 73.87  & 58.29 & 72.86 & 55.78 \\

    \midrule

    \multirow{6}{*}{InternVL 3.5}
        & Forward     & 100.00 & 100.00 & 100.00 & 100.00 \\
        & Backward    & 0.50   & 0.50   & 0.00   & 0.00 \\
        & Accelerate  & 99.50  & 97.49  & 99.50 & 95.48 \\
        & Decelerate  & 6.53   & 15.08  & 4.52 & 12.56 \\
        & Left        & 40.70  & 13.57  & 35.18 & 9.55 \\
        & Right       & 99.50  & 100.00 & 100.00 & 100.00 \\

    \midrule

    \multirow{6}{*}{MiniCPM-V 4.5}
        & Forward     & 100.00 & 100.00 & 100.00 & 100.00 \\
        & Backward    & 0.00   & 0.00   & 0.00   & 0.00 \\
        & Accelerate  & 31.16  & 0.50   & 28.14 & 0.00 \\
        & Decelerate  & 80.40  & 100.00 & 83.42 & 100.00  \\
        & Left        & 4.02   & 0.00   & 1.01 & 0.00 \\
        & Right       & 99.50  & 100.00 & 100.00 & 100.00 \\

    \bottomrule
    \end{tabular}}
    \label{tab:fewshot_methods_all_models}
\end{table}

% \begin{figure}[t]
%     \centering
%     \includegraphics[width=\columnwidth]{figures/qwen2_few_shot.png}
%     \caption{Comparison between few-shot prompting strategies for Qwen2-VL-7B-Instruct across ego-motion actions \textbf{TODO: MAKE THIS TYPE OF FIGURE ACROSS MODELS}.}
%     \label{fig:qwen_fewshot_methods}
% \end{figure}

% \begin{table}[t]
%     \centering
%     \footnotesize
%     \caption{Comparison between few-shot methods for VideoLLaMA2-7B. \textbf{Do we want to keep this?}}
%     \vspace{-1mm}
%     \resizebox{\columnwidth}{!}{%
%     \begin{tabular}{c || c c c c}
%     \toprule
%     Action 
%         & Few shot 
%         & Few shot with CoT 
%         & Few shot with VP 
%         & Few shot with CoT and VP \\
%     \midrule
%     Forward     & 100.0 & 100.0 & 100.0 & 100.0 \\
%     Backward    & 0.00  & 0.00  & 0.00  & 0.00 \\
%     Accelerate  & 100.00 & 100.00 & 100.0 & 100.0 \\
%     Decelerate  & 0.00  & 0.00  & 0.00  & 0.00 \\
%     Left        & 0.00  & 0.00  & 0.00  & 0.00 \\
%     Right       & 100.0 & 100.00 & 100.0 & 100.0 \\
%     \bottomrule
%     \end{tabular}}
%     \label{tab:videollama2_fewshot_methods}
% \end{table}

\vspace{-2mm}
\subsection{LoRA Finetuning}
\vspace{-1mm}
After the limitations of the prompt level approaches, we look to various learning techniques: starting with LoRA finetuning. Note that we employ 400 data points for each action pair, for example 200 forward and 200 backward, with an 80/20 train/held-out evaluation split. For all LoRA experiments, we train for 3 epochs with batch size 1 and learning rate $1\times10^{-4}$, using LoRA rank $r=16$, scaling factor $\alpha=32$, and dropout $0.05$. Following the process of finetuning, we are able to overcome the model biases seen before and learn the behaviors as seen in Tables \ref{tab:all_models_strategies} and \ref{tab:videollama2_strategies}. Here, we observe significantly higher levels of accuracy and show that at some level, these models are capable of learning the motions (with a slightly less performant VideoLLaMA2 for accelerate/decelerate) despite failing at a prompt level. In fact, besides the one aforementioned case, all models achieve near perfect accuracy with finetuning. \textbf{Thus, the failure is not architectural capacity, but accessibility: ego-motion cues are learnable under lightweight supervision, but not reliably exposed through prompting.} However, this LoRA approach converts the task into a supervised binary classifier and removes the open-ended reasoning language produced by the prompt-level methods. This motivates our feature-extraction analysis, which probes where in the visual pipeline motion-relevant information is encoded and why it fails to transfer to prompt-level spatio-temporal reasoning.
% \begin{table}[t]
%     \centering
%     \footnotesize
%     \caption{Aggregated performance across motion categories following LoRA finetuning for each model.}
%     \scalebox{0.9}{
%     \begin{tabular}{c || c c}
%     \toprule
%     Category & VideoLLaMA2 & Qwen2-VL \\
%     \midrule
%     Forward/Backward  & 100.00 & 100.0 \\
%     Accelerate/Decelerate    & 82.50  & 100.0 \\
%     Left/Right & 97.50  & 96.25 \\
%     \bottomrule
%     \end{tabular}}
%     \label{tab:lora_aggregated_motion_categories}
% \end{table}

\vspace{-2mm}
\subsection{Feature Extraction}
\vspace{-1mm}
We next probe whether ego-motion information is present in intermediate visual representations by training lightweight classifiers on extracted features. For VideoLLaMA2, we compare the vision-tower output $(8,576,1024)$ with the post-STC representation $(1,845,4096)$. For Qwen2-VL, Qwen2.5-VL, and Qwen3-VL, we probe immediately before and after visual token merging and projection; for InternVL 3.5, before and after visual downsampling/projection; and for MiniCPM-V 4.5, before and after its visual resampler. We use an 80/20 train/held-out evaluation split and train one-layer CNNs, with and without pooling, for up to 20 epochs.

As shown in Tables~\ref{tab:all_models_strategies}, \ref{tab:videollama2_strategies}, and~\ref{tab:cnn_features_all_models}, feature-level probes substantially outperform prompted inference, indicating that ego-motion information is encoded internally even when it is not reliably expressed by the full model. However, how this information evolves through the visual pipeline varies by architecture. In VideoLLaMA2, performance often decreases after the STC connector, while Qwen2.5-VL and Qwen3-VL improve markedly after visual merging: forward/backward rises from 50.00\% to 97.50\% and from 55.00\% to 97.50\%, respectively, and left/right reaches 100\% post-merge for both. MiniCPM-V 4.5 similarly improves from 67.50\% to 100\% on forward/backward, whereas InternVL 3.5 is already highly informative pre-merge, reaching 100\% on forward/backward and left/right and 98.75\% on accelerate/decelerate.

\textbf{Thus, motion information is not universally lost at a particular visual stage.} Depending on architecture, later processing may preserve, strengthen, or partially degrade it. The consistent finding is instead a gap between strongly decodable internal representations and the motion information reliably used by the final prompted prediction pathway.

% \begin{table}[t]
%     \centering
%     \footnotesize
%     \caption{Training simple linear probes based on the outputs of components of VideoLLaMA2.}
%     \resizebox{\columnwidth}{!}{%
%     \begin{tabular}{c || c c || c c}
%     \toprule
%     \multirow{2}{*}{Action Pair}
%         & \multicolumn{2}{c||}{Linear probe}
%         & \multicolumn{2}{c}{Linear probe + pooling} \\
%     & Vision Tower & STC Connector
%     & Vision Tower & STC Connector \\
%     \midrule
%     Forw/Back  & 58.75 & 60.00 & 60.00 & 57.50 \\
%     Acc/Dec    & 58.75 & 73.75 & 62.50 & 71.25 \\
%     Left/Right & 93.75 & 90.00 & 97.50 & 90.00 \\
%     \bottomrule
%     \end{tabular}}
%     \label{tab:linear_feat_vidllama}
% \end{table}

% \begin{table}[t]
%     \centering
%     \footnotesize
%     \caption{Training simple CNNs based on the outputs of components of VideoLLaMA2.}
%     \resizebox{\columnwidth}{!}{%
%     \begin{tabular}{c || c c || c c}
%     \toprule
%     \multirow{2}{*}{Action Pair}
%         & \multicolumn{2}{c||}{1-layer CNN}
%         & \multicolumn{2}{c}{1-layer CNN + pooling} \\
%     & Vision Tower & STC Connector
%     & Vision Tower & STC Connector \\
%     \midrule
%     Forw/Back  & 98.75  & 77.50  & 100.00 & 66.25 \\
%     Acc/Dec    & 86.25  & 66.25  & 82.50  & 62.50 \\
%     Left/Right & 100.00 & 100.00 & 100.00 & 96.25 \\
%     \bottomrule
%     \end{tabular}}
%     \label{tab:cnn_feat_vidllama}
% \end{table}

\begin{table}[t]
    \vspace{-3mm}
    \centering
    \caption{Training simple CNN probes on intermediate representations from different video-language models. Results are reported for features extracted before and after the visual feature compression stage. $^\dagger$ Chance-level pooled-probe run.}
    \tiny
    \resizebox{0.775\columnwidth}{!}{%
    \begin{tabular}{c || c || c c || c c}
    \toprule
    \multirow{2}{*}{Model}
        & \multirow{2}{*}{Action Pair}
        & \multicolumn{2}{c||}{1-layer CNN}
        & \multicolumn{2}{c}{1-layer CNN + pooling} \\
    & & Pre & Post
      & Pre & Post \\
    \midrule

    \multirow{3}{*}{VideoLLaMA2}
        & Forw/Back  & 98.75  & 77.50  & 100.00 & 66.25 \\
        & Acc/Dec    & 86.25  & 66.25  & 82.50  & 62.50 \\
        & Left/Right & 100.00 & 100.00 & 100.00 & 96.25 \\

    \midrule

    \multirow{3}{*}{Qwen2-VL}
        & Forw/Back  & 88.75 & 93.75 & 88.75 & 92.50 \\
        & Acc/Dec    & 87.50 & 88.75 & 87.50 & 86.25 \\
        & Left/Right & 93.75 & 97.50 & 95.00 & 97.50 \\

    \midrule

    \multirow{3}{*}{Qwen2.5-VL}
        & Forw/Back  & 50.00 & 97.50  & 50.00$^\dagger$ & 50.00$^\dagger$ \\
        & Acc/Dec    & 83.75 & 91.25  & 85.00 & 50.00$^\dagger$ \\
        & Left/Right & 77.50 & 100.00 & 50.00$^\dagger$ & 50.00$^\dagger$ \\

    \midrule

    \multirow{3}{*}{Qwen3-VL}
        & Forw/Back  & 55.00 & 97.50 & 50.00$^\dagger$ & 50.00$^\dagger$ \\
        & Acc/Dec    & 83.75 & 85.00 & 50.00$^\dagger$ & 50.00$^\dagger$ \\
        & Left/Right & 52.50 & 100.00 & 50.00$^\dagger$ & 50.00$^\dagger$ \\

    \midrule

    \multirow{3}{*}{InternVL 3.5}
        & Forw/Back  & 100.00 & 65.00  & 97.50 & 96.25 \\
        & Acc/Dec    & 98.75  & 93.75  & 95.00 & 95.00 \\
        & Left/Right & 100.00 & 100.00 & 100.00 & 100.00 \\

    \midrule

    \multirow{3}{*}{MiniCPM-V 4.5}
        & Forw/Back  & 67.50  & 100.00 & 51.25 & 50.00$^\dagger$ \\
        & Acc/Dec    & 97.50  & 97.50  & 97.50 & 97.50 \\
        & Left/Right & 100.00 & 100.00 & 98.75 & 50.00$^\dagger$ \\

    \bottomrule
    \end{tabular}}
    \label{tab:cnn_features_all_models}
\end{table}

% \begin{table}[t]
%     \centering
%     \caption{Accuracy of 1-layer CNN probes trained on visual features extracted from temporally shuffled videos. Frames are shuffled before being passed through each video-language model, disrupting their original temporal ordering while preserving the set of observed frames.}
%     \tiny
%     \resizebox{0.75\columnwidth}{!}{%
%     \begin{tabular}{c || c c c}
%     \toprule
%     Model
%         & Forw/Back
%         & Left/Right
%         & Acc/Dec \\
%     \midrule
%     Qwen2-VL
%         & 51.25
%         & 100.00
%         & 90.00 \\
%     \midrule
%     Qwen2.5-VL
%         & 61.25
%         & 98.75
%         & 82.50 \\
%     \midrule
%     Qwen3-VL
%         & 66.25
%         & 100.00
%         & 88.75 \\
%     \midrule
%     InternVL 3.5
%         & 53.75
%         & 100.00
%         & 97.50 \\
%     \midrule
%     MiniCPM-V 4.5
%         & 76.25
%         & 100.00
%         & 100.00 \\
%     \bottomrule
%     \end{tabular}}
%     \label{tab:shuffled_feature_probes}
% \end{table}

\paragraph{Summary of Findings.}
\begin{itemize}
\item Prompting-based performance is highly model- and task-dependent. While few-shot prompting improves selected motion categories for newer models, strong class biases and near-chance performance remain common, particularly for forward/backward motion.
\item Lightweight LoRA finetuning achieves near-perfect performance, showing that the architectures can readily learn ego-motion distinctions under supervision.
\item Representation-level probing reveals that motion-relevant information is strongly decodable within the visual hierarchy, although its evolution across processing stages is architecture-dependent.
\end{itemize}

\textbf{In summary, these results support our central claim: current video-language models encode and can learn ego-motion information, but do not reliably exploit it during prompted inference. In the driving domain, their off-the-shelf predictions therefore often behave more like frame-level, bias-driven decisions than temporal reasoning.}

\vspace{-2mm}
\section{Conclusion}
\vspace{-2mm}

We studied whether modern video-language models reason over temporal dynamics in driving videos or instead behave primarily as frame-based recognizers. Across VideoLLaMA2, Qwen2-VL, Qwen2.5-VL, Qwen3-VL, InternVL 3.5, and MiniCPM-V 4.5, standard prompting reveals strong and often systematic class biases, including complete collapse on the forward/backward task for all evaluated models. These behaviors also persist in our smaller real-world evaluation, where standard-prompting class-level performance is strongly correlated with CARLA ($r=0.957$). More structured prompting does not consistently resolve these failures: few-shot demonstrations produce substantial gains for selected newer models and tasks, but these improvements do not transfer consistently across motion categories.

At the same time, these failures do not imply that motion information is absent from the models. LoRA finetuning achieves substantially higher accuracy, often near-perfect, showing that the architectures can learn motion-sensitive behavior under lightweight supervision. Feature extraction further shows that simple probes can recover motion cues from intermediate visual representations, even when the full prompted model fails. However, because we do not include single-frame or frame-shuffling controls, high LoRA and probe accuracy cannot by itself establish that these cues are genuinely temporal as some performance may reflect per-frame appearance cues in the CARLA renders and this is left to future work. How this information evolves across the visual stack varies by architecture, but the common finding is a gap between what is encoded internally and what is reliably used by the final prompted prediction pathway. Thus, the central failure is not simply a lack of visual information, but a failure to reliably exploit motion-relevant cues for prediction.

Overall, our results caution against interpreting fluent video-language outputs as evidence of dependable temporal reasoning in driving. Current models can encode and learn ego-motion cues, and prompting can partially expose these capabilities, but off-the-shelf temporal predictions remain unstable. Future work should test whether these findings persist on more expansive real-world driving datasets such as nuScenes and the Waymo Open Dataset and explore architectures or training objectives that more directly encourage temporal information (including Liquid structural state-space models~\cite{hasani2022liquid}, selective state-space models such as Mamba~\cite{gu2023mamba}, and Linear Oscillatory State-Space models~\cite{rusch2025oscillatory}) to be preserved and utilized.

% \vspace{-2mm}
% \section*{ACKNOWLEDGMENT}
% \vspace{-1mm}
% Research was sponsored by the Department of the Air Force Artificial Intelligence Accelerator and was accomplished under Cooperative Agreement Number FA8750-19-2-1000. The views and conclusions contained in this document are those of the authors and should not be interpreted as representing the official policies, either expressed or implied, of the Department of the Air Force or the U.S. Government. The U.S. Government is authorized to reproduce and distribute reprints for Government purposes notwithstanding any copyright notation herein. Author Alaa Maalouf thanks the Neubauer Family Foundation and the Council for Higher Education's MAOF program for their support.

% \section*{References}

%
% ---- Bibliography ----
%
\vspace{-3mm}
\bibliographystyle{IEEEtran}
\bibliography{references}

\end{document}